\documentclass{article}
\usepackage{spconf,amsmath,graphicx}

\usepackage{enumitem}
\setlist{nosep, leftmargin=14pt}

\usepackage{mwe} 
\usepackage{amsmath}
\usepackage{amssymb}
\usepackage{enumitem}
\usepackage{lipsum}
\usepackage{cite}
\usepackage{comment}
\usepackage{physics}
\usepackage{bm}

\newcommand{\Tref}[1]{Table~\ref{#1}}
\newcommand{\Fref}[1]{Fig.~\ref{#1}}
\newcommand{\Eref}[1]{Eq.~\ref{#1}}

\title{Ranking-Guided Semi-Supervised Domain Adaptation for Severity Classification}
\name{Shota Harada$^{\star}$ \qquad Ryoma Bise$^{\star}$ \qquad Kiyohito Tanaka$^{\dagger}$ \qquad Seiichi Uchida$^{\star}$}
\address{$^{\star}$ Kyushu University, Japan \\
$^{\dagger}$ Kyoto Second Red Cross Hospital, Japan}

\begin{document}
\maketitle
\begin{abstract}
Semi-supervised domain adaptation leverages a few labeled and many unlabeled target samples, making it promising for addressing domain shifts in medical image analysis. However, existing methods struggle with severity classification due to unclear class boundaries. Severity classification involves naturally ordered class labels, complicating adaptation. We propose a novel method that aligns source and target domains using rank scores learned via ranking with class order. Specifically, Cross-Domain Ranking ranks sample pairs across domains, while Continuous Distribution Alignment aligns rank score distributions. Experiments on ulcerative colitis and diabetic retinopathy classification validate the effectiveness of our approach, demonstrating successful alignment of class-specific rank score distributions.
\end{abstract}
\section{Introduction}
\label{sec:intro}
Domain shift problems, arising from differences between training and test data, pose significant challenges in medical image analysis. For example, a disease classifier trained on labeled images from one hospital~(source) often underperforms on data from a different hospital~(target) due to variations in imaging equipment, conditions, and protocols. These differences lead to discrepancies in feature distributions between source and target domains, impairing the classifier’s generalization ability. Therefore, effective domain adaptation methods are crucial for improving classifier performance across different medical settings.
\par
Semi-supervised domain adaptation (SSDA) effectively addresses domain shift by leveraging a few labeled and many unlabeled target samples. SSDA methods can improve performance over unsupervised domain adaptation with a few efforts, making them appealing for constructing diagnostic support systems. However, SSDA for severity classification is challenging due to the ambiguity of class boundaries caused by the continuity of classes, where classes represent ordered levels of disease severity rather than distinct classes.
\par
Most existing domain adaptation methods assume that classes form distinct clusters in the feature space, which enables alignment across domains by matching class-specific clusters. However, this assumption does not hold for severity classification, where class boundaries are inherently blurred, and distinct clusters for each class are difficult to achieve. Consequently, applying traditional SSDA methods for common classification~\cite{He2020cvprws,Pin2020ijcai,li2021cross,li2023adaptive,li2024inter,li2021ECACL,saito2019semi,Yu_2023_CVPR} to this task often leads to suboptimal results, as these methods fail to capture the ordered nature of severity levels, resulting in misaligned feature distributions.
\par
\begin{figure*}[t]
    \centering
    \includegraphics[width=.8\textwidth]{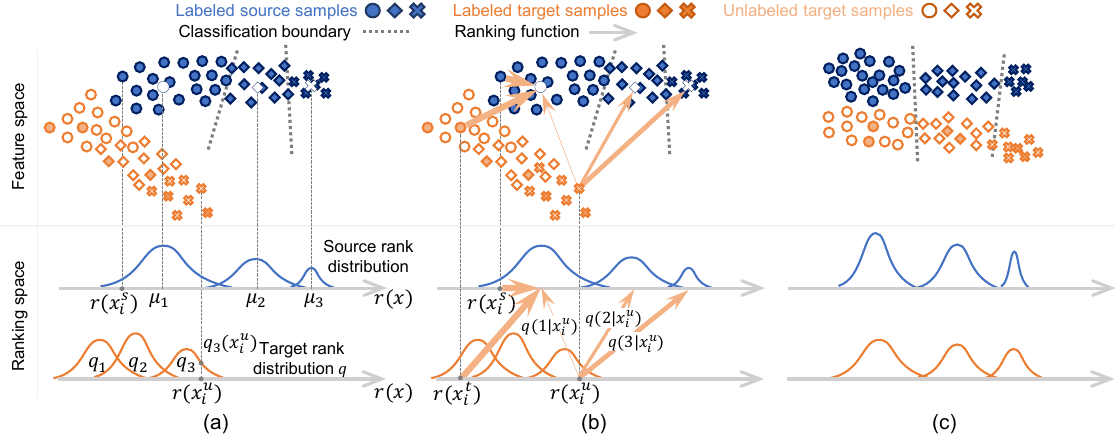}
    \vspace{-3mm}
    \caption{Overview of the proposed method. The top and bottom rows show the feature and ranking space, respectively. The rank score is derived by applying the ranking function $r(\bm{x})$ to the features in the top row. (a) Initial distribution in a semi-supervised domain adaptation scenario, considering class order ($\times \succ \diamond \succ \circ$), and target rank distribution estimation for unlabeled target samples. (b) Association of source and target rank distributions with rank-based soft labels, where arrow size indicates the association weight. (c) Semi-supervised domain adaptation using cross-domain ranking and continuous distribution alignment.}
    \label{fig:overall}
\end{figure*}
To address this issue, we propose a novel adaptation approach that leverages a rank score representing continuous severity progression obtained through learning to rank. Learning to rank estimates which sample in a pair is more severe based on its relative label, which represents the severity order based on their class labels. We first apply learning to rank~\cite{ranknet} to the source samples, learning a continuous severity-based feature distribution and ranks (classes), as shown in~\Fref{fig:overall}~(a). When applied to the target samples, the feature distribution deviates from the source distribution due to domain shift, as shown at the bottom of \Fref{fig:overall}~(a). As a result, the rank score distribution for each class shifts. Here, similar to many other domain adaptation methods that assume feature vectors of the same class form a cluster, we assume that, despite the shift in the target distribution, samples within each class retain a similar severity-based ordering.
\par
In this study, we align the feature distributions by matching the rank score distributions of the shared ranking function. Aligning the distributions as discrete classes is ineffective due to the ambiguous class boundaries caused by class continuity. In contrast, by treating the classes as rank scores and aligning those scores, we mitigate class continuity. This approach introduces the continuous progression of severity as rank scores into the SSDA process, enabling domain adaptation that takes into account the fact that, in reality, each sample belongs to a continuous severity rather than a discrete class. Specifically, this is achieved using the following components: 
\begin{enumerate}
    \item Cross-Domain Ranking (CDR): CDR performs ranking using relative severity labels between source and target samples, ensuring consistent rank scores across domains. By aligning samples based on relative ranking, CDR establishes a common severity scale, ensuring consistent severity interpretation. Unlike traditional ranking, which operates within a single domain, CDR extends ranking across domains for meaningful correspondence.
    
    \item Continuous Distribution Alignment (CDA): CDA refines alignment by estimating rank-based soft labels, leveraging the class continuity of unlabeled target samples. Unlike hard pseudo-labels, these labels probabilistically associate samples with classes, capturing continuous severity. CDA minimizes the distance between source and target rank score distributions, ensuring effective domain alignment despite ambiguous class boundaries.
\end{enumerate}
\par
CDR and CDA together align rank scores between domains, thereby aligning feature distributions based on class order. By constraining feature distributions to reflect the ordinal structure of severity levels, the proposed method enables effective domain adaptation for severity classification, where clear class distinctions are often absent. Aligning rank score distributions through ordered class constraints allows for domain-independent feature extraction and improves performance in these tasks for medical imaging.
\par
Our contributions are summarized as follows:
\begin{itemize}
\item We propose a ranking-guided SSDA framework for severity classification that aligns source and target domains using ordered class labels, addressing ambiguous class boundaries.
\item We introduce CDA with the rank-based soft label, which is a simple and effective loss function for domain adaptation on this task. Additionally, we introduce CDR, a novel approach for SSDA in this task that learns rankings from cross-domain sample pairs.
\end{itemize}
\par
\section{Ranking-Guided Semi-Supervised Domain Adaptation}
\label{sec:prop}
In semi-supervised domain adaptation of severity classification, we are given $N_s$ labeled source samples $\mathcal{D}^s=\{(\bm{x}_i^s, y_i^s )\}_{i=1}^{N_s}$, where $\bm{x}_i^s$ is the $i-$th sample in the source domain and $y_{i}^{s}\in\{1, ..., C\}$ is its ordered class label. We are also given $N_t$ labeled target samples $\mathcal{D}^t=\{(\bm{x}_i^t, y_i^t )\}_{i=1}^{N_t}$ and $N_u$ unlabeled target samples $\mathcal{D}^u=\{\bm{x}_i^u\}_{i=1}^{N_u}$. We aim to improve the classification performance in the target domain utilizing target samples $\mathcal{D}^t$ and $\mathcal{D}^u$ and source samples $\mathcal{D}^s$.
\par
The network in the proposed method is a hierarchical structure in which a feature extractor~$F_e$ branches into a classifier~$F_c$ and a ranking function~$F_r$. A sample $\bm{x}_i$ is passed through $F_e$ to obtain the features, which are then passesd through $F_c$ and $F_r$ to obtain the class probability $\bm{p}(\bm{x}_i)=F_c(F_e(\bm{x}_i))\in[0, 1]^C$ and the rank score $r(\bm{x}_i)=F_r(F_e(\bm{x}_i))\in\mathbb{R}$, respectively.
\par
The proposed method is optimized in two steps: first, using $\mathcal{D}^s$, it is optimized with pairwise ranking loss, which uses relative labels derived from class labels as ground-truth, and classification loss; second, it is further optimized by cross-domain ranking on $\mathcal{D}^s$ and $\mathcal{D}^t$, along with continuous distribution alignment using $\mathcal{D}^s$, $\mathcal{D}^t$, and $\mathcal{D}^s$.
\par
\noindent{\bf Cross-Domain Ranking}:\
\label{sec:cdr}
To address domain shifts in the rank scores of labeled source and target samples, we propose cross-domain ranking (CDR), which trains pairwise rankings from pairs sampled across domains. While class labels are typically treated independently in each domain, we mitigate the domain shift by learning to rank across domains. We pair labeled source and target samples $\mathcal{D}^s \cup \mathcal{D}^t$ for ranking, ensuring consistent rankings across domains and aligning the rank score distributions of both domains.
\par
For the CDR, we input a pair of labeled samples, $\bm{x}_i$ and $\bm{x}_j$, sampled from $\mathcal{D}^s\cup\mathcal{D}^t$ into $F_e$, and then minimize the ranking loss~\cite{ranknet} which is defined as:

{
\footnotesize
\begin{align}
\label{eq:rank}
\mathcal{L}_{\textrm{r}}(\bm{x}_i, \bm{x}_j)&=-o_{i,j}\log h \left(r_{i,j}\right)-(1-o_{i,j})\log\left(1- h \left(r_{i,j}\right)\right), \\
o_{i,j}&= \nonumber
\begin{cases}
    \bm{1}(y_{i}>y_{j}) & \textrm{if}~y_i \neq y_j, \\
    \frac{1}{2} & \textrm{if}~y_i = y_j,
\end{cases}
\end{align}
}
where $h(\cdot)$ is a sigmoid function, and $r_{i,j}$ is $r(\bm{x}_i)-r(\bm{x}_j)$. This loss function forces samples with larger classes to have higher rank scores than those with smaller classes. Along with CDR, we minimize \Eref{eq:rank} within the same domain and the cross-entropy loss $\mathcal{L}_{c}$ using $\mathcal{D}^s$ and $\mathcal{D}^t$ for classification.
\par
\begin{figure}[t]
    \centering
    \includegraphics[width=\columnwidth]{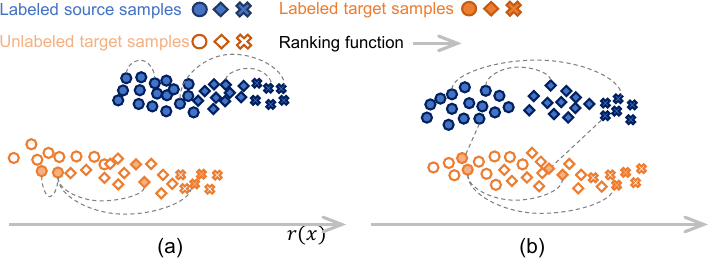}
    \vspace{-8mm}
    \caption{Effect of Cross-Domain Ranking (CDR). (a) Distribution when learning to rank within the same domain. (b) Distribution with CDR. The dashed lines represent sample pairs.}
    \vspace{-6mm}
    \label{fig:cdr}
\end{figure}
As shown in \Fref{fig:cdr}, CDR is essential in semi-supervised domain adaptation. Ranking with a single domain improves class discrimination within each domain, but the class distributions may not align across domains. For example, in \Fref{fig:cdr}~(a), the rank score of the $\diamond$ class sample in the target is smaller than that of the $\circ$ class sample in the source (with the actual order being $\diamond \succ \circ$). In contrast, CDR avoids such discrepancies, as shown in \Fref{fig:cdr}~(b), aligning the rank score distributions between domains. Thus, cross-domain ranking is crucial in semi-supervised domain adaptation.
\par
\noindent{\bf Continuous Distribution Alignment}:\
\label{sec:rda}
Continuous Distribution Alignment (CDA) aligns the rank score distributions of both domains by leveraging associations between samples and the rank score distribution in the source domain, as shown in \Fref{fig:overall}~(b). CDA uses rank-based soft labels, derived from the estimated rank score distribution of unlabeled target samples, to associate them with the source domain's rank score distributions. Labeled samples from both domains are similarly associated with the source distribution. CDA then minimizes the distance between the rank scores and their corresponding distributions, aligning rank score distributions and, ultimately, the feature distributions across domains.
\par
CDA trains in the following steps. We first model the rank score distribution for each class of unlabeled target samples $\mathcal{D}^u$ using GMM with the number of cluster $C$, which is the same as the number of classes, and obtain the $k$-th cluster probability $q(k|\bm{x}_i^u)$. Next, we minimize the CDA loss to reduce the distance between the rank score of each sample and the average rank score for each class in the source domain $\mu_k$ as a class prototype, which is defined as:

{
\footnotesize
\begin{align}
\label{eq:rda}
\mathcal{L}_a(\bm{x}_i)&=\sum_{k=1}^{C}{w_{i,k}} \left(r(\bm{x}_i)-\mu_k\right)^2, \\
\mu_k &= \frac{1}{\sum_{y_n^s\in\mathcal{D}^s}\bm{1}(y_{n}^s = k)} \sum_{(\bm{x}_j^s, y_j^s)\in\mathcal{D}^s} \bm{1}(y_{j}^s = k) r(\bm{x}_j^s),\nonumber\\
w_{i,k}&= \nonumber
\begin{cases}
    \bm{1}(y_{i}=k) & \textrm{if}~\bm{x}_i \in \mathcal{D}^s\cup\mathcal{D}^t, \\
    q(k|\bm{x}_i) & \textrm{if}~\bm{x}_i \in \mathcal{D}^u,
\end{cases}
\end{align}
}
where $w_{i,k}$ is the rank-based soft label that indicates the assignment ratio for class $k$ of $\bm{x}_i$. The labeled samples are directly assigned to their class, while unlabeled samples are assigned based on rank-based soft labels, maintaining their original rank order while matching the source distribution.
\par
The overall loss function is defined as follows:

{
\footnotesize
\begin{align}
\label{eq:total}
\mathcal{L} &= \sum_{\bm{x}_i, \bm{x}_j \in\mathcal{D}^s\cup\mathcal{D}^t} \{\mathcal{L}_c(\bm{x}_i)+ \mathcal{L}_c(\bm{x}_j) +\mathcal{L}_r(\bm{x}_i, \bm{x}_j)\}\nonumber\\
&+\lambda \sum_{\bm{x}_l\in\mathcal{D}^s\cup\mathcal{D}^t\cup\mathcal{D}^u}\mathcal{L}_a(\bm{x}_l),
\end{align}
}
where $\lambda$ is a hyper-parameter to balance CDA and the others. 
\section{Experiments}
\label{sec:exp}
\begin{table*}[t]
    \centering
    \caption{Classification results for target test samples when the number of labeled target samples per class is $10$. \textbf{Bold} indicates the best, while \underline{underline} indicates the second best, and mP, mR, and mF1 are the mean of the macro Precision, the macro Recall, and the macro F1, respectively.}
    \vspace{-3mm}

\label{tab:perf}
    \centering
    \begin{tabular}[t]{l c c c c| c c c c}
    \hline
    Method & Accuracy & mP & mR & mF1 & Accuracy & mP & mR & mF1\\
     & \multicolumn{4}{c|}{LIMUC$\rightarrow$Private} & \multicolumn{4}{c}{DDR$\rightarrow$FGADR}\\
    \hline
    S & $0.369$ & $0.352$ & $0.481$ & $0.288$ & $0.323$ & $0.437$ & $0.354$ & $0.248$\\
    S+T & $0.582$ & $0.444$ & $0.580$ & $0.444$ & $\underline{0.422}$ & $0.427$ & $0.408$ & $\underline{0.367}$\\
    MME~\cite{saito2019semi} & $\mathbf{0.613}$ & $\underline{0.470}$ & $\mathbf{0.596}$ & $\underline{0.477}$ & $0.420$ & $0.392$ & $\underline{0.434}$ & $0.356$\\
    CDAC~\cite{li2021cross} & $0.599$ & $0.429$ & $0.553$ & $0.430$ & $0.404$ & $0.397$ & $0.405$ & $0.351$\\
    SLA~\cite{Yu_2023_CVPR} & $0.540$ & $0.448$ & $0.578$ & $0.441$ & $0.383$ & $\underline{0.452}$ & $0.379$ & $0.310$\\
    ORUDA~\cite{chidlovskii2021universal} & $0.555$ & $0.433$ & $0.531$ & $0.441$ & $0.342$ & $0.337$ & $0.331$ & $0.289$\\
    Ours & $\underline{0.611}$ & $\mathbf{0.521}$ & $\underline{0.587}$ & $\mathbf{0.494}$ & $\mathbf{0.470}$ & $\mathbf{0.470}$ & $\mathbf{0.454}$ & $\mathbf{0.407}$\\
    \hline
\end{tabular}
\end{table*}

\noindent{\bf Dataset}:\
In this experiment, we used two classification tasks: ulcerative colitis~(UC) classification of endoscopic images and diabetic retinopathy~(DR) classification of fundus images. For each task, we prepared two datasets. For UC, we used LIMUC~\cite{limuc} and a private dataset collected from Kyoto Second Red Cross Hospital~(Private), both of which are annotated by experts using a 4-level Mayo score, one of the severity scores for UC. The number of images in LIMUC and Private were $11{,}276$ and $10{,}265$, respectively. For DR, we used FGADR~\cite{fgadr} and DDR~\cite{ddr}, both of which are annotated by experts on a 5-level scale. The number of images in FGADR and DDR was $1{,}842$ and $11{,}017$, respectively.
\par
We used LIMUC and DDR as the source, the rest as the target. We randomly split the source domain into patient units, with $60\%$, $20\%$, and $20\%$ of the samples allocated to training, validation, and testing, respectively. The validation set was used only during pretraining. To perform 5-fold cross-validation, we applied patient-disjoint random sampling to the target. During training, we randomly selected $10$ samples per class as labeled target samples $\mathcal{D}^t$, and $10$ samples per class for validation.
\par
\noindent{\bf Comparative Methods}:\
We compared the proposed method with the six methods: \textbf{S}, trained with only labeled source samples; \textbf{S+T}, trained with both labeled source and labeled target samples; \textbf{MME}~\cite{saito2019semi}, \textbf{CDAC}~\cite{li2021cross}, and \textbf{SLA}~\cite{Yu_2023_CVPR}, semi-supervised domain adaptation methods; \textbf{ORUDA}~\cite{chidlovskii2021universal}, unsupervised domain adaptation method for ordinal classification. For a fair comparison, \textbf{ORUDA} used the labeled target samples $\mathcal{D}_t$ to learn the same loss function as $\mathcal{D}_s$.
\par
\noindent{\bf Implementation Details}:\
ResNet-50~\cite{kaiming2016CVPR} pretrained on ImageNet~\cite{olga2015ijcv} was used as a backbone network for the feature extractor $F_e$ with its last linear layer removed. A linear layer was used for both the classifier $F_c$ and the ranking function $F_r$. During pretraining and adaptation, we applied class-weighted random sampling to obtain mini-batches with approximately an equal number of labeled samples per class. To train with standard ranking loss and the CDR, we randomly combined samples in a mini-batch to make sample pairs. The parameter $\lambda$ in \Eref{eq:total} was set to $10^{-5}$ and $10^{-6}$ for the UC classification and the DR classification. We terminated the training of the model by early stopping, referred to as the macro F1 (mF1) of the validation set.
\par
\noindent{\bf Experimental Results}:\
\Tref{tab:perf} shows the classification performance of the target test samples in the UC and DR tasks with $10$ labeled target samples per class. This confirmed that the proposed method outperformed the others in mP and mF1. However, several methods outperformed it in mR. Given the class imbalance in the dataset, mF1 is the most crucial metric, and the proposed method achieved the best performance. 
\par
\Tref{tab:ablation} shows the results of the ablation study with CDR and CDA. The results demonstrate that both CDR and CDA effectively improve mP and mR. In particular, CDR contributes to the improvement of mP, while CDA enhances mR, which is a reasonable trend since CDA aligns the entire data distribution between both domains, while CDR relies on labeled samples. The significant improvement when both are trained simultaneously suggests they work complementarily.
\par
\begin{table}[t]
\centering
\caption{The ablation study with Cross-Domain Ranking~(CDR) and Continuous Distribution Alignment~(CDA) in LIMUC$\rightarrow$Private.}
    \vspace{-3mm}

\label{tab:ablation}
\begin{tabular}[t]{l l | c c c c}
    \hline
    CDR & CDA & Accuracy & mP & mR & mF1 \\
    \hline
     & & $0.588$ & $0.455$ & $0.572$ & $0.447$ \\
    \checkmark & & $\mathbf{0.615}$ & $0.479$ & $0.588$ & $0.484$ \\
    & \checkmark & $0.606$ & $0.478$ & $\mathbf{0.589}$ & $0.472$ \\
    \checkmark & \checkmark & $0.611$ & $\mathbf{0.521}$ & $0.587$ & $\mathbf{0.494}$ \\
    \hline
\end{tabular}
\vspace{-3mm}
\end{table}
\par
\begin{figure}[t]
    \centering
    \includegraphics[width=\columnwidth]{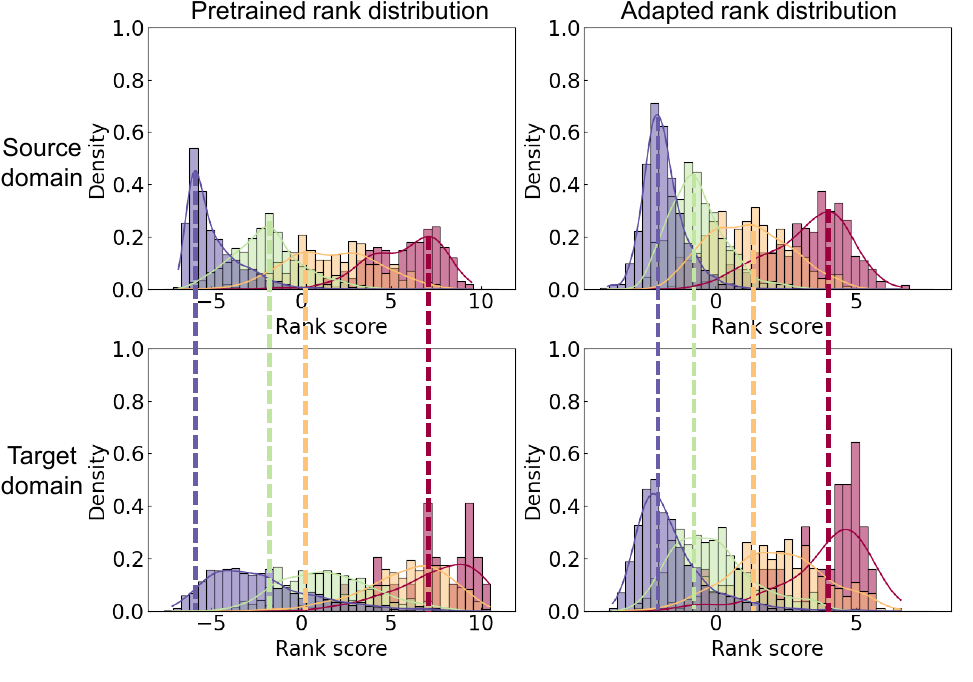}
    \vspace{-10mm}
    \caption{Visualization of the rank score distribution in LIMUC$\rightarrow$Private. The horizontal and vertical axes represent the rank score and sample density, respectively. 
    }
    \vspace{-3mm}
    \label{fig:k2l_rank}
\end{figure}

\Fref{fig:k2l_rank} shows the rank score distributions before and after adaptation. The horizontal and vertical axes represent the rank score and sample density, respectively. The top and bottom rows show the source and target domains, respectively, with color indicating class and dashed lines showing the source mode for each class. Before adaptation, the distributions do not match between domains, but after applying the proposed method, the distributions are aligned for each domain and class, demonstrating its effectiveness in this task.
\par

\section{Conclusion}
\label{sec:conc}
In this paper, we proposed a semi-supervised domain adaptation method for severity classification leveraging learning to rank for continuous class boundaries. It consists of Cross-Domain Ranking, which aligns domain distributions by ranking labeled sample pairs made from cross-domains, and Continuous Distribution Alignment, which aligns rank score distributions between domains with rank-based soft labels. Experiments on two severity classification tasks demonstrated its effectiveness. The experimental results confirmed that the model pretrained on the source domain aligns target samples according to class order, and that the adaptation step enhances class discrimination while preserving class alignment. 
\par
\clearpage
\section{Compliance with ethical standards}
This study was performed in line with the principles of the Declaration of Helsinki. Approval was granted by the Ethics Committee of Kyoto Second Red Cross Hospital.

\section{Acknowledgments}
\label{sec:acknowledgments}
This work was supported by JST ACT-X Grant Number JPMJAX23CN, JST ASPIRE Grant Number JPMJAP2403, and AMED Grant Number JP25mk0121326.

\bibliographystyle{IEEEbib}
\bibliography{strings,myref}

\end{document}